\documentclass[conference]{IEEEtran}
\usepackage{cite}
\usepackage[pdftex]{graphicx}
\usepackage{amsmath}
\usepackage{array}
\usepackage{tabularx}
\usepackage{multirow}

\begin{document}

\title{MultiSiam: A Multiple Input Siamese Network For Social Media Text Classification And Duplicate Text Detection}

\author{
\IEEEauthorblockN{
Sudhanshu Bhoi\IEEEauthorrefmark{1},
Swapnil Markhedkar\IEEEauthorrefmark{2},
Shruti Phadke\IEEEauthorrefmark{3}, and
Prashant Agrawal\IEEEauthorrefmark{4}}
\IEEEauthorblockA{
Department of Computer Engineering\\
Pune Institute of Computer Technology,\\
Pune, India}
\IEEEauthorblockA{Email: \{
\IEEEauthorrefmark{1}sudh.bhoi.30,
\IEEEauthorrefmark{2}swapnilmarkhedkar,
\IEEEauthorrefmark{3}shrutidphadke,
\IEEEauthorrefmark{4}prashantsa255
\}@gmail.com}}

\maketitle

\begin{abstract}
Social media accounts post increasingly similar content, creating a chaotic experience across platforms, which makes accessing desired information difficult. These posts can be organized by categorizing and grouping duplicates across social handles and accounts. There can be more than one duplicate of a post, however, a conventional Siamese neural network only considers a pair of inputs for duplicate text detection. In this paper, we first propose a multiple-input Siamese network, MultiSiam. This condensed network is then used to propose another model, SMCD (Social Media Classification and Duplication Model) to perform both duplicate text grouping and categorization. The MultiSiam network, just like the Siamese, can be used in multiple applications by changing the sub-network appropriately.
\end{abstract}

\IEEEpeerreviewmaketitle

\section{Introduction}
A significant contributor to the millions of bytes of data produced daily are the myriad of social media networks in use today \cite{usageStats}. A sizable proportion of these social media posts are comparable, or duplicates. This, combined with the increasingly popular infinite feed format makes accessing desirable information difficult. We aim to optimize the user’s feed in two ways: first, by dividing posts into different categories based on their content, and second, by aggregating duplicate posts on and across different social media into one post.

The two natural language processing problems, categorization, and duplicate detection involve some common steps. Duplicate posts will belong to the same category, which means word embeddings calculated for categorization can be used for duplicate detection. Developing a common model which can handle both the tasks and understand the relation between them, will avoid repetitive calculations and improve the application pipeline.

For a user, more than 2 duplicate posts can be found across different handles or various social media accounts. This requires more than a pair of posts to be compared. Conventional Siamese models only consider a pair of inputs, which means, to compare groups of inputs, pairs need to be prepared through careful batching to ensure no two pairs from the same batch belong to the same group. Moreover, making all possible pairs from a group makes learning redundant. 

In this paper, we propose a modified Siamese neural network, MultiSiam, with a condensed architecture for duplicate detection, which can handle multiple inputs at a time. We further use this to create a model, SMCD (Social Media Classification and Duplication Model), for the proposed application of social media text classification and duplicate text detection

\section{Background and Related Work}
The Siamese network, first introduced in \cite{siameseFirst}, has been successfully implemented across computer vision \cite{computerVision} and natural language processing \cite{nlpAppln} applications. Various distance measures, like Euclidean distance \cite{euclidD}, Jaccard coefficient \cite{jaccardD}, Cosine similarity \cite{cosine} have been used to find similarity between 2 outputs of the sub-networks. This paper \cite{tripletLoss} uses Cosine similarity and the widely used Triplet loss. Other losses include Contrastive loss \cite{contrastiveLoss}, one of the oldest, which considers the distance between 2 inputs, while Quadruplet loss \cite{quadLoss} considers 4 inputs. In 2020, \cite{circleLoss} introduced Circle loss which aims to maximize within-class similarity and minimize between-class similarity.

Text categorization is a classical problem in NLP. \cite{textCat} reviews more than 150 deep learning models including those based on RNN, CNN, and the attention mechanism. For duplicate question-pair detection, the Siamese network has out-performed multi-layer perceptron architecture and the traditional shingling algorithm \cite{dupQue}. Siamese can also be used for text categorization \cite{siamText}. It is especially helpful when the number of categories is large, not known during training and examples per category are small \cite{cvSimilarity}. 

Social media related tasks can be improved by considering different modalities, text and image \cite{multimodal}, hyperlinks \cite{links}, and hashtags \cite{tweetCat} from a post.

\section{Network Architecture}
\subsection{Multiple Input Siamese Network (MultiSiam)}
\label{section:multiSiam}
A Siamese network has 2 input fields with 2 separate sub-networks to act on them \cite{siameseFirst}. Many applications like signature verification \cite{siameseFirst} or duplicate question detection \cite{dqd} require the sub-networks to be identical, having the same parameters and weights. This paper proposes a network that can accept 2 or more input fields with a sub-network to act on them.

\begin{figure}[!t]
    \centering
    \includegraphics[width=0.9\linewidth]{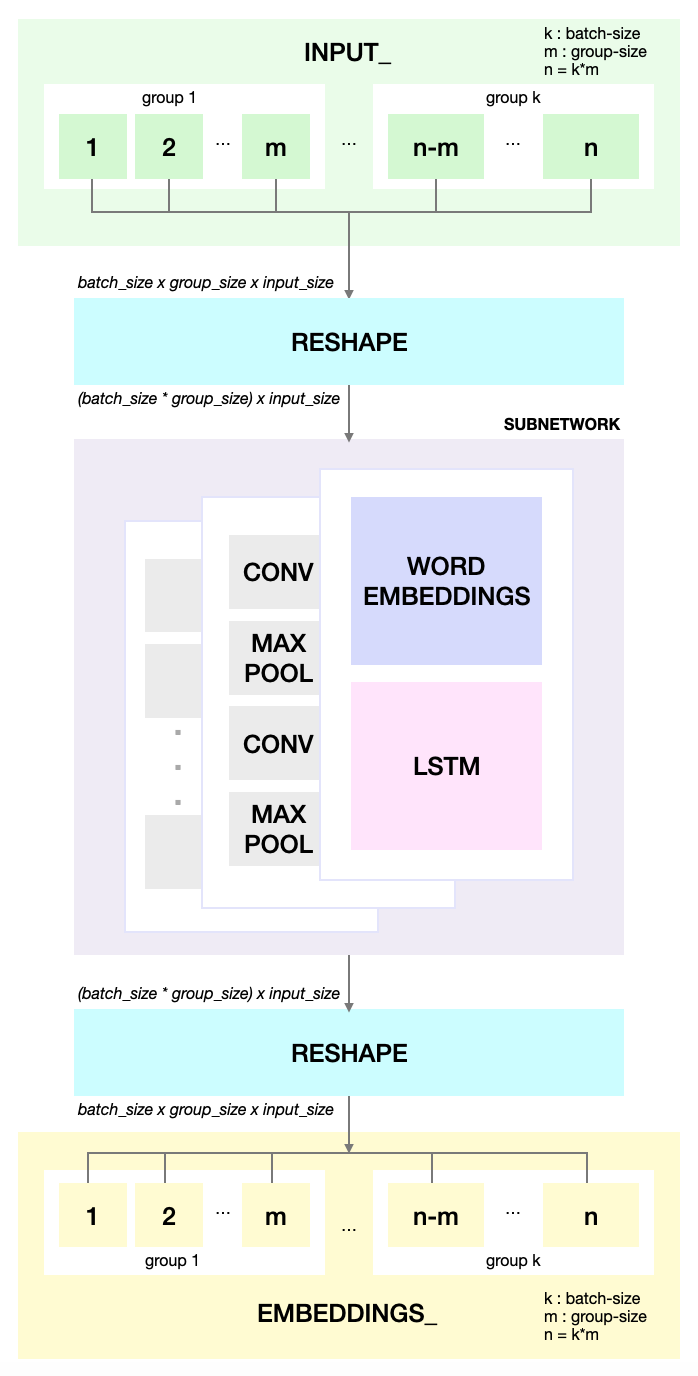}
    \caption{The MultiSiam network architecture with example sub-networks.}
    \label{fig:multisiamArch}
\end{figure}

The MultiSiam network, shown in Figure \ref{fig:multisiamArch}, expects a batch of groups of input fields. In case of duplicate detection, a group contains all texts that are duplicates of each other. Each batch should have groups of equal length. Each group should have texts of equal length. Text can be padded with zeros or a special token from vocabulary, while a group can be padded by sampling a required number of texts from the same group. Thus the input has dimension of $batch\_size$ x $group\_size$ x $text\_size$.

This input is then reshaped to $(batch\_size * group\_size)$ x $text\_size$ for training. It is then passed through a sub-network. In case of duplicate detection, the sub-network consists of a Word Embedding layer and an LSTM layer \cite{dqd}. Note that this can be replaced with any subnetwork based on the application as shown in Figure \ref{fig:multisiamArch}.  

The output has dimension of $(batch\_size * group\_size)$ x $embedding\_size$. This is finally reshaped into $batch\_size$ x $group\_size$ x $embedding\_size$. Thus the network produces embeddings for each of the input fields. These embeddings are then used for calculating the triplet loss \cite{tripletLoss}.

\subsubsection{Triplet Loss}
The triplet loss aims to minimize the distance between the anchor and positive and maximizes the distance between the anchor and negative \cite{tripletLoss}. From the network output, if one text is considered as an anchor then all the other texts in the same group are positives while those belonging to other groups are negatives. This can be represented as a matrix as explained in \cite{siamDeepAI} for $group\_size$ of 2. Here we generalize it to dimension $batch\_size_1$ x $batch\_size_2$ x $...$ x  $batch\_size_{group\_size}$. 

Each cell in the matrix represents the cosine similarity of the corresponding embeddings. The diagonal of such a matrix represents the cosine similarity with positives, while off-diagonal represents that with negatives. Considering the first in each group to be the anchor, the mean of all the negatives corresponding to it and the highest off-diagonal value is calculated, resulting in 2 $batch\_size$ vectors $mean\_neg$ and $closest\_neg$ respectively \cite{siamDeepAI}. 

Triplet Loss is then calculated as follows:
\begin{equation}
    dist_i^{ap} = cos(f(x_i^a),f(x_i^p))
\end{equation}
\begin{equation}
    dist_{ij}^{an} = cos(f(x_i^a),f(x_{ij}^n))_{j \neq i}
\end{equation}
\begin{equation}
    mean\_neg_i = (\sum_j^{N-1} dist_{ij}^{an})_{j \neq i} / (N-1)
\end{equation}
\begin{equation}
    closest\_neg_i = max_j(dist_{ij}^{an})_{j \neq i}
\end{equation}
\begin{equation}
    cost_i^1 = max(-dist_i^{ap} + mean\_neg_i + \alpha, 0)
\end{equation}
\begin{equation}
    cost_i^2 = max(-dist_i^{ap} + closest\_neg_i + \alpha, 0)
\end{equation}
\begin{equation}
    cost_i = cost_i^1 + cost_i^2
\end{equation}
\begin{equation}
    TripletLoss = \sum_{i}^{N} cost_i
\end{equation}
Where $x^a, x^p, x^n$ are anchor, positive, and negative inputs respectively. $f(x)$ is the embedding from the network. $\alpha$ is the margin between positive and negative pairs. $N$ is the $batch\_size$.

\subsubsection{Inference}
\begin{figure*}[!t]
    \centering
    \includegraphics[width=0.8\textwidth]{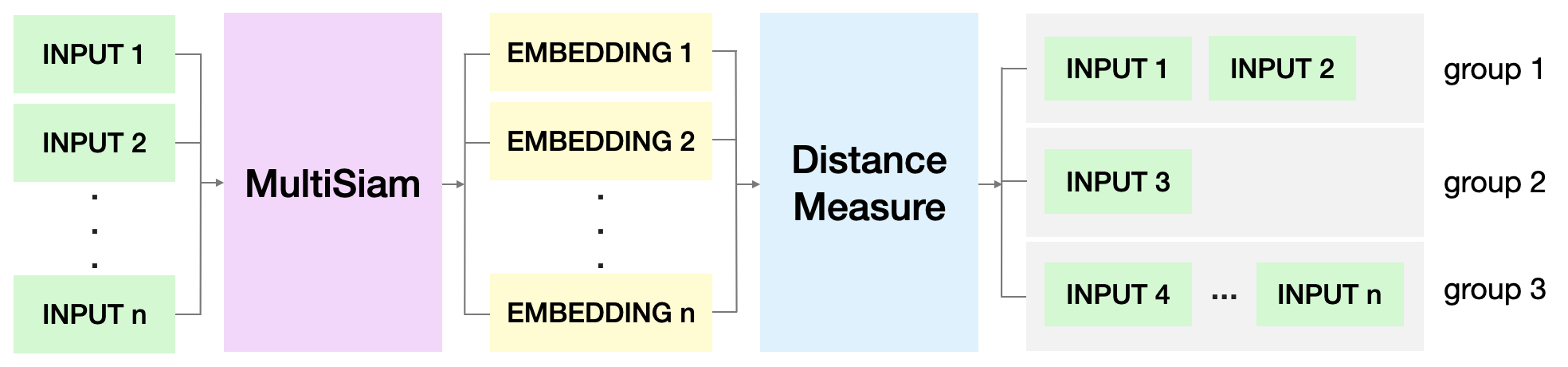}
    \caption{MultiSiam, during inference, with distance measure and example groups of given inputs based on the embeddings produced.}
    \label{fig:multisiamDist}
\end{figure*}

Inference can be of two types. One, groups need to be formed from given inputs, shown in Figure \ref{fig:multisiamDist}. In this, the Reshape layer is disabled so the MultiSiam produces embeddings for the inputs. This is similar to using only one of the sub-networks in Siamese during evaluation. Distance measures like cosine similarity and a threshold value can be used between the embeddings to group inputs together. Two, already formed groups need to be verified. In this, MultiSiam outputs groups of embeddings as in Figure \ref{fig:multisiamArch}. A distance measure between embeddings of a group and a threshold value then can determine whether the group constitutes  duplicates. 

\subsection{Social Media Classification and Duplication Model (SMCD)}
In the proposed application, duplicate texts are likely to be found in a single category. So a single embedding can be used for categorization as well as further processed for duplicate detection. This model leverages the above fact to combine the two tasks using the concise MultiSiam network. 

\begin{figure}[!t]
    \includegraphics[width=\linewidth]{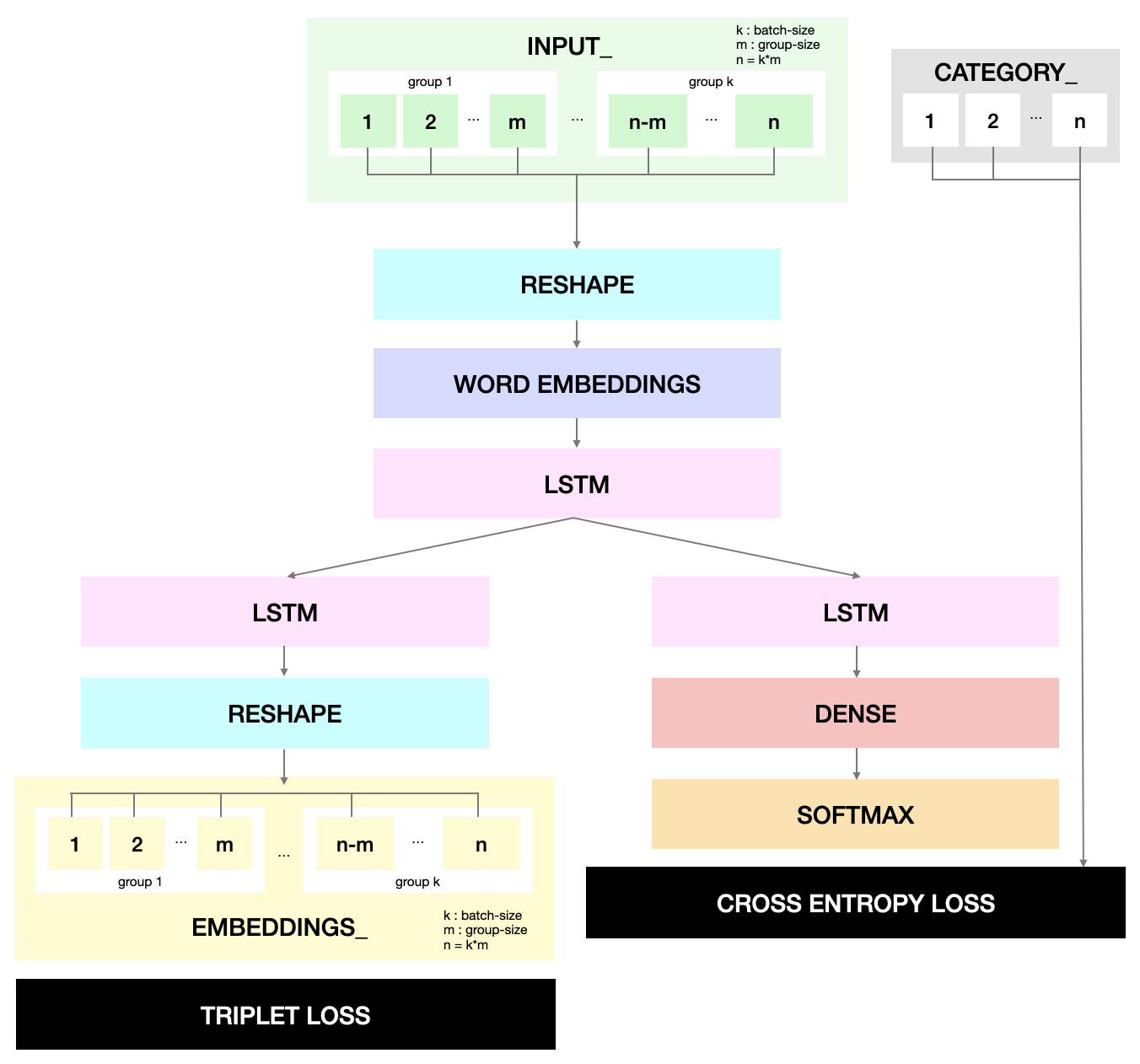}
    \caption{SMCD model for text categorization and duplicate text detection with respective losses.}
    \label{fig:smcdArch}
\end{figure}

First, the input is reshaped according to the MultiSiam network. Then it is passed to the Word Embeddings and LSTM layers to produce embeddings E. These embeddings, E, are then passed through an LSTM layer, a Dense layer, and a Softmax function to get a category. The same embeddings, E, are also passed to another LSTM layer \cite{lstm} and then reshaped (as in the MultiSiam) to get the embeddings for duplicate detection. Refer Figure \ref{fig:smcdArch}.

Cross-entropy loss is calculated on predicted and actual categories of dimension $(batch\_size * group\_size)$ x $num\_categories$. Triplet loss is calculated on the final embeddings as mentioned in the MultiSiam above. Thus, for a given batch of texts, this proposed model finds their duplicate embeddings as well as their categories. A distance measure, like cosine similarity, can be used between these embeddings to get groups of duplicate texts as shown in Figure \ref{fig:multisiamDist}.

\section{Experiments and Results}
\label{section:results}
The experiments were performed using TensorFlow v2.5 on a 2.3 GHz Quad-Core Intel Core i5 CPU. These experiments aimed to show the similarity between proposed MultiSiam and existing Siamese for a pair of inputs under the same conditions. They also demonstrate the model's usefulness with more than 2 inputs in a group and integration with the categorization task in the SMCD model.

\subsection{MultiSiam}
We trained a MultiSiam and a Siamese model on the Quora duplicate questions dataset \cite{quoraDataset}. The dataset consists of 404348 pairs of questions with a corresponding label indicating whether they are duplicates. Out of these 149306 are duplicate pairs and were considered for training with a 90:10 split. Further 14931 non-duplicates pairs were added to the test set. Each model was then trained for 10 epochs with batches of 256 examples. Triplet Loss had a margin of 0.25 and a pair was considered duplicate if its cosine similarity was above 0.7. Although MultiSiam is a little slower due to dimensional operations, it performs very similarly to the Siamese model in terms of accuracy. Refer Table \ref{table:modelComp}.

\begin{table}[!t]
    \centering
    \caption{Model Comparison}
    \label{table:modelComp}
    \begin{tabular}{|l|c|c|}
    \hline
    \textbf{Model} & \textbf{Accuracy} & \textbf{CPU Time}\\
    \hline
    Siamese & 0.74 & 1 s/step \\
    \hline
    MultiSiam & 0.72 & 3 s/step \\
    \hline
    \end{tabular}
\end{table}

For more than 2 inputs, experiments were performed on a custom dataset in the SMCD model.

\subsection{SMCD}
We selected 13 different categories to classify posts into, namely: Business, Politics, Celebrity, Entertainment, Finance, Health \& Wellness, Motivation, Sports, News, Promotions, Gaming, Travel, and Technology.  

\begin{table*}[!t]
    \centering
    \caption{Sample from the custom dataset prepared to train the SMCD model before any processing}
    \label{table:smcdDataset}
    \begin{tabularx}{\textwidth}[h!]{|X|l|c|}
        \hline
        \textbf{Text} & \textbf{Category} & \textbf{Group ID}\\
        \hline
        Person A officially takes over as Company A CEO from Person B & Business & 1 \\
        \hline
        Today is Day 1 for Person A as former Company B CEO moves to Company A corner office URL by MENTION & Business & 1 \\
        \hline
        Person A takes over the role of Company A CEO from Person B as the company faces an antitrust push. MENTION reports. & Business & 1 \\
        \hline
        Company B Software A beta is here and ready for you to use & Tech & 2 \\
        \hline
        Software A public beta hands-on: These 5 features make all the difference & Tech & 2 \\
        \hline
        Here’s what’s new in Software A beta 2 & Tech & 2 \\
        \hline
        The best new features and changes in Software A beta 2 & Tech & 2 \\
        \hline
    \end{tabularx}
\end{table*}

To create the dataset, we sourced posts from 10-12 social media handles per category (from Reddit, Facebook, and Twitter) which were determined to exclusively post content belonging to that particular category, and labeled them accordingly. The labeling errors, if any, were corrected manually. Duplicate posts were sourced by querying multiple handles for the same content topic, both within and across the three social media networks.

The dataset thus collected had around 15 groups per category. Each group consisted of, on average, 2 to 4 texts from the posts. Table \ref{table:smcdDataset} shows a sample of the dataset (anonymized for this paper). As mentioned in section \ref{section:multiSiam}, these groups were padded or truncated to have a group size of 4. Thus the dataset had 201 groups in total and 804 texts after processing. Grouped texts were used to train the duplicate-detection-half of the model while corresponding grouped category labels were used to train the categorization-half. The model was trained for 20 epochs with a batch size of 16.

During the evaluation, we provided a list of texts for which we needed duplicate embeddings and the category the inputs belonged to. These embeddings were then used to group the inputs. Table \ref{table:smcdOut} demonstrates an output from the model.

\begin{table}[!t]
    \centering
    \caption{Duplicate Grouped and Categorized texts from the SMCD model output obtained by providing list of texts as input}
    \label{table:smcdOut}
    \begin{tabularx}{\linewidth}{|c|X|l|}
    \hline
    \textbf{Group ID} & \textbf{Texts} & \textbf{Categories}\\
    \hline
    \multirow{3}{*}{1} & Former President announces law suit against Company A & News \\ \cline{2-3}
    & Lawsuit announced against Company A by President Person A & News \\ \cline{2-3}
    & Former President plans to sue Company A & News \\
    \hline
    \multirow{2}{*}{2} & TV Show A predicted Pandemic A outbreak in 1993? & Entertainment  \\ \cline{2-3}
    & Fans of ’TV Show A’ convinced show predicted the pandemic & Entertainment \\
    \hline
    \end{tabularx}
\end{table}

\subsection{Application}
The application \cite{proRepo}, for every user, aggregates their feed (across different social media networks) into one optimized feed, with posts separated into different categories, and duplicate posts identified and clubbed into one. This is done using the SMCD model proposed above.

\section{Conclusion and Future Scope}
The accuracy in section \ref{section:results} can be improved by various techniques like transfer learning, text pre-processing, hyper-parameter tuning, etc. Further, a standard multi-input duplicate dataset can be created for various tasks to evaluate the models. Different loss functions, sub-networks can be also experimented with in future work.

Currently, with the help of the proposed MultiSiam network, we were able to create a condensed model, SMCD, that could categorize texts and group duplicate texts together. In the future, with NLP techniques like text summarization, social media feeds can be made more organized. We believe such a system will help improve users’ mental health as well as increase their productivity.

The MultiSiam network can also be used for computer vision, natural language processing, and various other tasks by using an appropriate sub-network. It brings all the goodness of Siamese networks to more than 2 inputs at a time.

\section*{Acknowledgment}
The authors would like to thank Prof. R. V. Bidwe and Vedang Mandhana for their continued guidance and support throughout the research and development process.

\end{document}